\documentclass[sigconf]{acmart}

\graphicspath{{./Figures/}}
\DeclareGraphicsExtensions{.pdf}

\usepackage{graphicx}
\usepackage{amsmath} 
\usepackage{algorithm}
\usepackage{algorithmic}
\usepackage{multirow}
\usepackage{color}
\usepackage{subfigure}
\usepackage{enumitem}
\setlist{leftmargin=2.5ex}

\copyrightyear{2024}
\acmYear{2024}
\setcopyright{acmlicensed}\acmConference[DAC '24]{61st ACM/IEEE Design Automation Conference}{June 23--27, 2024}{San Francisco, CA, USA}
\acmBooktitle{61st ACM/IEEE Design Automation Conference (DAC '24), June 23--27, 2024, San Francisco, CA, USA}
\acmDOI{10.1145/3649329.3656538}
\acmISBN{979-8-4007-0601-1/24/06}

\begin{document}

\title{
\vspace{-1.5ex}
FeBiM: Efficient and Compact Bayesian Inference Engine Empowered with Ferroelectric In-Memory Computing}

 \author{\small{}
    \vspace{-1.5ex}
    Chao Li$^1$, 
    Zhicheng Xu$^2$,
    Bo Wen$^2$,
    Ruibin Mao$^2$,
    Can Li$^2$, 
    Thomas K{\"a}mpfe$^3$,
    Kai Ni$^4$ and
    Xunzhao Yin$^{1,5*}$
    \\$^1$College of Information Science and Electronic Engineering, Zhejiang University
    \\$^2$Department of Electrical and Electronic Engineering, The University of Hong Kong; 
    $^3$Center Nanoelectric Technologies, Fraunhofer IPMS
    \\$^4$Electrical Engineering Department, University of Notre Dame; 
    $^5$Key Laboratory of CS\&AUS of Zhejiang Province
    \\\vspace{-1.5ex}$^*$Corresponding author email: xzyin1@zju.edu.cn
    \vspace{-1ex}
}

\renewcommand{\bibfont}{\scriptsize}
\let\oldbibliography\thebibliography
\renewcommand{\thebibliography}[1]{\oldbibliography{#1}
\setlength{\itemsep}{-0.5pt}}

\begin{abstract}
\vspace{-1ex}
In scenarios with limited training data or where explainability is crucial, conventional neural network-based machine learning models often face challenges.
In contrast, Bayesian inference-based algorithms excel in providing interpretable predictions and reliable uncertainty estimation in these scenarios.
While many state-of-the-art in-memory computing (IMC) architectures leverage emerging non-volatile memory (NVM) technologies to offer unparalleled computing capacity and energy efficiency for neural network workloads, their application in Bayesian inference is limited.
This is because the core operations in Bayesian inference, i.e., cumulative multiplications of prior and likelihood probabilities, differ significantly from the multiplication-accumulation (MAC) operations common in neural networks, rendering them generally unsuitable for direct implementation in most existing IMC designs.
In this paper, we propose FeBiM, an efficient and compact Bayesian inference engine powered by multi-bit ferroelectric field-effect transistor (FeFET)-based IMC.
FeBiM effectively encodes the trained probabilities of a Bayesian inference model within a compact FeFET-based crossbar.
It maps quantized logarithmic probabilities to discrete FeFET states.
As a result, the accumulated outputs of the crossbar naturally represent the posterior probabilities, i.e., the Bayesian inference model's output given a set of observations.
This approach enables efficient in-memory Bayesian inference without the need for additional calculation circuitry. 
As the first FeFET-based in-memory Bayesian inference engine, FeBiM achieves an impressive storage density of 26.32 Mb/mm$^{2}$  and a computing efficiency of 581.40 TOPS/W  in a representative Bayesian classification task.
These results demonstrate 10.7$\times$/43.4$\times$ improvement in compactness/efficiency compared to the state-of-the-art hardware implementation of Bayesian inference.

\end{abstract}

\maketitle
\pagestyle{empty}

\vspace{-1.5ex}
\section{Introduction}
\vspace{-1ex}
\label{sec:intro}

In-memory computing (IMC) has recently emerged as a promising solution to address the memory wall issues in conventional von Neumann hardware \cite{ielmini2018memory}.
Leveraging the compactness and high energy efficiency of emerging non-volatile memory (NVM) technologies, many leading IMC accelerators have achieved impressive computing efficiency and throughput for data-intensive machine learning models, particularly neural networks (NNs) \cite{shafiee2016isaac, hu2021memory, yan2023improving, jung2022crossbar}.
While conventional NN-based algorithms are widely used, they often struggle in situations where training data is insufficient or when interpretable results are needed \cite{qayyum2020secure, yang2022unbox}.
As a compelling alternative, Bayesian inference is particularly well-suited in low-data scenarios, providing explainable results with uncertainty estimation \cite{ghahramani2015probabilistic, burkart2021survey}.

The primary posterior calculation in Bayesian inference, which involves the cumulative product of prior and likelihood probabilities as per Bayes' theorem, however, differs from the multiply-and-accumulate (MAC) operations common in NN workloads.
This difference renders Bayesian inference usually unsuitable for direct implementation with many existing IMC designs that typically focus on NN acceleration.
In traditional complementary metal-oxide-semiconductor (CMOS)-based von Neumann implementations for Bayesian inference, such as CPU \cite{smith2020massively}, GPU \cite{talbot2019parallelized} and field-programmable gate array (FPGA) \cite{awano2020bynqnet}, accessing separate memory units for stored probabilities incurs significant area and energy overhead.
Efforts to exploit the non-volatility and energy efficiency of emerging devices have led to the development of Bayesian inference prototypes utilizing random number generators (RNGs) built with magnetic tunnel junction (MTJ) \cite{vodenicarevic2017low}, memtransistor \cite{zheng2022hardware} and magnetic random-access memory (MRAM) \cite{faria2018implementing}.
These implementations, however, are limited to Bayesian inference with binary evidence/events, and do not effectively address probability storage, rather generating required probabilities on demand, which is energy-consuming.
A memristor-based Bayesian machine has been proposed \cite{harabi2023memristor}, using near-memory stochastic computing to reduce memory access overhead.
Yet, these implementations still require additional CMOS logic and multiple clock cycles for posterior calculations and complex sensing circuitry for final inference, thus compromising computing density and inference efficiency. 

To address the aforementioned challenges in hardware implementation of Bayesian inference, we propose FeBiM, an efficient and compact in-memory Bayesian inference engine utilizing multi-level cell (MLC) ferroelectric field-effect transistors (FeFETs). 
The key contributions of this work are summarized as follows:
\vspace{-1.5ex}
\begin{itemize}
  \item We propose a compact crossbar array design using one FeFET per cell as probability storage unit and a compact and scalable winner-take-all (WTA) circuit for sensing.
  This multi-bit FeFET array enables efficient in-memory Bayesian inference in just one clock cycle, eliminating the need for extra calculation circuitry. 
  \item We introduce a novel mapping scheme that associates quantized logarithmic probabilities with discrete FeFET states.
  This scheme enables the output currents of the crossbar to naturally represent the posterior probabilities, i.e., the cumulative product of priors and likelihoods given a set of observations.
  \item We thoroughly investigate the functionality, scalability and application level performance of FeBiM.
  In a representative Bayesian classification task, our proposed design shows a 10.7$\times$/43.4$\times$ storage density/inference efficiency improvement compared to the state-of-the-art Bayesian machine.
\end{itemize}
\vspace{-1ex}

The rest of the paper is organized as follows.
Section \ref{sec:background} reviews the basics and relevant prior works. 
Section \ref{sec:design} introduces our FeFET-based IMC design for Bayesian inference. 
Section \ref{sec:eval} presents the validation, scalability investigation and application benchmarking results of FeBiM.
Finally, section \ref{sec:conclusion} concludes the paper.

\vspace{-1.5ex}
\section{Preliminaries and Related Works}
\label{sec:background}
\vspace{-1ex}

In this section, we briefly introduce the basics of FeFET device and its advantages in IMC, as well as Bayesian inference preliminaries and existing hardware implementations.

\vspace{-1.5ex}
\subsection{In-Memory Computing with FeFET}
\label{subsec:fefet}
\vspace{-1ex}

\begin{figure}[!t]
\vspace{-1.5ex}
\centering
\includegraphics[width=0.94\columnwidth]{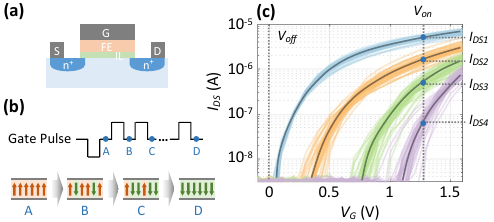}
\vspace{-3ex}
\caption{
(a) Schematic of a FeFET device.
(b) Partial polarization switching in an MFM capacitor induced by a write pulse train.
(c) The multi-level $I_{D}$-$V_{G}$ characteristics.
Each dark curve represents a FeFET state. 
}
\vspace{-4ex}
\label{fig:fefet}
\end{figure}

Recent years have seen significant research interests in FeFET.
Unlike two-terminal resistive devices, the three-terminal FeFET functions as a standalone 1T non-volatile storage unit, featuring a high $ON/OFF$ ratio, high energy efficiency, small footprint and CMOS compatibility \cite{muller2012ferroelectricity, khan2020future, yin2024ferroelectric}.
As shown in Fig.\ref{fig:fefet}(a), a FeFET integrates HfO$_2$ as the ferroelectric dielectric layer within the gate stack of a MOSFET \cite{dunkel2017fefet}.
Write operations of FeFETs involve applying positive/negative voltage pulses to the gate. 
This switches the polarization in the ferroelectric layer towards the channel/gate metal, setting the FeFET into a low-$V_{TH}$/high-$V_{TH}$ state.
The polarization switching in FeFETs, driven by the electric field, results in superior write energy efficiency ($\sim$fJ/bit)  compared to other NVM devices \cite{ni2019ferroelectric}.
The binary states of FeFETs have been extensively studied and applied in existing IMC designs \cite{yin2022ferroelectric}.

By varying the number of positive write pulses applied to the gate of FeFET after a full erase, as shown in Fig.\ref{fig:fefet}(b), partial polarization switching and multiple distinct $V_{TH}$ states can be realized.
For instance, in a 2-bit storage case shown in Fig.\ref{fig:fefet} (c), 4 $V_{TH}$ states with well-controlled device variation are achieved.
When activated with $V_{on}$ applied to the gate, a FeFET shows a current $I_{DS}$ between the drain and source.
When inhibited with $V_{off}$, the FeFET is in a cut-off state.
The FeFET states are thus associated with corresponding $I_{DS}$ values.
Exploiting the multi-level characteristics of FeFET, recent IMC designs \cite{li2020scalable, shou2023see, soliman2023first, yin2023ultracompact, yin2021deep} have achieved improved density while maintaining computing robustness,  compared to the single-level cell (SLC)-based and analog approaches, respectively.

\vspace{-1.5ex}
\subsection{Bayesian Inference}
\label{subsec:bayes}
\vspace{-1ex}

Bayesian inference is a key probabilistic framework that facilitates decision-making in scenarios with incomplete information.
It incorporates evidence, assumptions, and prior knowledge to make informed decisions \cite{box2011bayesian}.
Unlike conventional NN-based machine learning approaches, Bayesian models excel at providing interpretable results and accurately estimating prediction certainty, even with limited data \cite{ghahramani2015probabilistic, burkart2021survey}.
Bayesian inference is widely applied in various fields, including medical diagnosis, where limited patient data requires the integration of expert knowledge \cite{nikovski2000constructing}, and in machine learning for decision-making under uncertainty \cite{trimmer2011decision}.

At its core, Bayesian inference leverages Bayes' theorem to update the probability of an event based on observed evidence:
\vspace{-1ex}
\begin{equation}
\label{equ:bayes}
    P(A|B) = \frac{P(A) \cdot P(B|A)}{P(B)}
\vspace{-1ex}
\end{equation}

Here, $P(A)$ is the prior probability of event $A$, indicating the initial belief or knowledge, and $P(B)$ is the probability of observing evidence $B$.
$P(B \lvert A)$ is the likelihood probability of observing $B$ given that event $A$ has occurred.
$P(A \lvert B)$ represents the updated posterior probability of event $A$ occurring, given that $B$ is observed.
Since the evidence probability $P(B)$ is constant for a given $A$, it is often omitted in calculations without affecting the order of magnitude of posteriors. Thus Eq. \eqref{equ:bayes} can be simplified as:
\vspace{-1ex}
\begin{equation}
\label{equ:bayes2}
    P(A|B_{1}, B_{2}, ..., B_{n}) \propto P(A) \cdot P(B_{1}, B_{2}, ..., B_{n}|A)
\vspace{-1ex}
\end{equation}
For ease of modeling and calculation, the likelihood factors $P(B_{i})$ can be considered conditionally independent in many real-life scenarios without compromising the prediction accuracy \cite{lowd2005naive}, and Eq. \eqref{equ:bayes2} can further be simplified as:
\vspace{-2ex}
\begin{equation}
\label{equ:bayes3}
    P(A|B) \propto P(A) \cdot \prod_{i=1}^{n} P(B_{i}|A)
\vspace{-1ex}
\end{equation}
In applications such as diagnosis/classification, the event with the highest posterior probability is selected as the final cause/category:
\vspace{-3ex}
\begin{equation}
\label{equ:bayes4}
    \hat{A} = \arg \max_{A} P(A|B) = \arg \max_{A} P(A) \cdot \prod_{i=1}^{n} P(B_{i}|A)
\vspace{-1ex}
\end{equation}

The cumulative product of prior and likelihoods in Eq. \eqref{equ:bayes3} is different from the commonly-used MAC operations in NN-based models.
As a result, many existing IMC designs dedicated to MAC acceleration do not directly support Bayesian inference.
Implementing  Bayesian inference in hardware requires addressing the storage of model parameters (i.e.,  priors and likelihoods), the posterior calculations (Eq. \eqref{equ:bayes3}) and final decision-making (Eq. \eqref{equ:bayes4}).
Conventionally, Bayesian inference models have been implemented on von Neumann platforms such as CPU \cite{smith2020massively}, GPU \cite{talbot2019parallelized}, FPGA \cite{awano2020bynqnet} and application-specific integrated circuit (ASIC) \cite{ko20203mm}.
However, these platforms face bottlenecks in frequent data transfer between external memory and computing units, severely affecting the throughput and efficiency.
To efficiently realize posterior calculations, some emerging NVM devices have been used to build RNGs in small-scale Bayesian inference prototypes, demonstrating compactness and better energy efficiency compared to traditional CMOS device \cite{vodenicarevic2017low, zheng2022hardware, faria2018implementing}.
However, these works do not address the storage and access of probabilities, a fundamental aspect of Bayesian inference implementation, and require dedicated circuits for on-demand probabilities generation.
To mitigate excessive memory access, \cite{harabi2023memristor} builds digital NVM with memristors that performs near-memory posterior calculations, greatly reducing data movement overhead compared to conventional implementations.
Yet, this approach requires additional CMOS logic and multiple clock cycles per posterior calculation and final sensing, leading to lower density and efficiency.

\vspace{-1.8ex}
\section{FeBiM Design and Workflow}
\vspace{-0.5ex}
\label{sec:design}

\begin{figure}[!t]
\centering
\vspace{-2ex}
\includegraphics[width=\columnwidth]{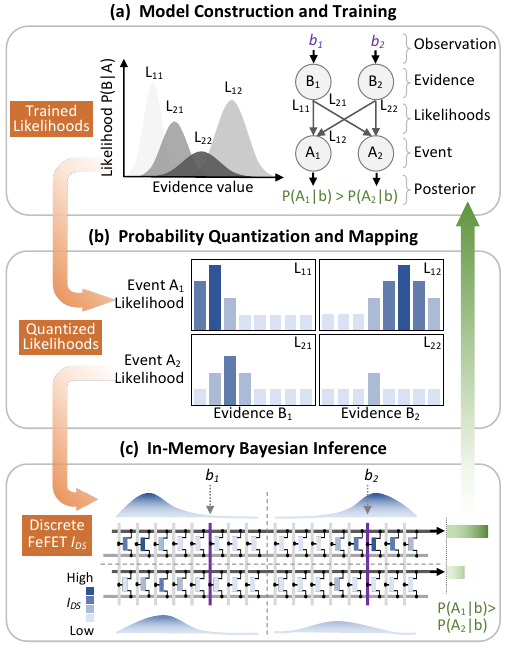}
\vspace{-6ex}
\caption{
The overall workflow of FeBiM.
Trained probabilities of the Bayesian model are quantized and mapped to discrete FeFET states.
Given observed evidence values, the FeFET-based crossbar outputs maximum posterior.
}
\label{fig:overview}
\vspace{-5ex}
\end{figure}

\begin{figure*}[!t]
\centering
\vspace{-2ex}
\includegraphics[width=0.95\textwidth]{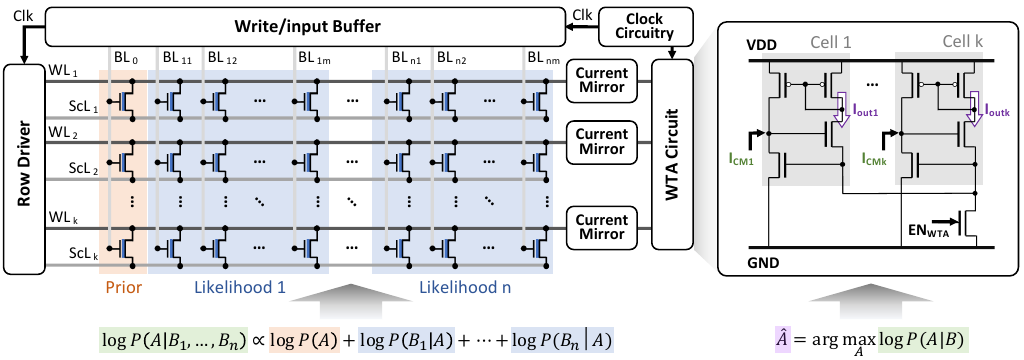}
\vspace{-3.5ex}
\caption{
The proposed crossbar array effectively programs the priors and likelihoods in multi-bit FeFET cells.
During inference, the posteriors are naturally calculated on each $WL$ and fed into the WTA circuit, which detects the maximum posterior.
}
\label{fig:array}
\vspace{-3.5ex}
\end{figure*}

To overcome the challenges in existing hardware implementation of Bayesian inference, we introduce FeBiM, a compact and efficient Bayesian inference engine empowered with multi-bit FeFET-based IMC.
In this section, we present the overall workflow of FeBiM, the core array design 
with just one multi-bit FeFET device per cell, and an effective scheme that maps the probabilities to the crossbar.

\begin{figure}[h]
\centering
\includegraphics[width=0.98\columnwidth]{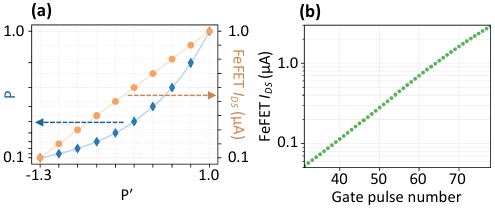}
\vspace{-3.2ex}
\caption{
(a) Truncated probabilities $P$ are first converted to the logarithmic values and normalized to $P^{\prime}$, which is then quantized and mapped to discrete FeFET states.
(b) Relation between the gate pulse number and the FeFET state. 
}
\label{fig:mapping}
\vspace{-4.5ex}
\end{figure}

\vspace{-2ex}
\subsection{Overview of FeBiM}
\label{subsec:overview}
\vspace{-0.5ex}

Fig.~\ref{fig:overview} illustrates the workflow of FeBiM.
To begin with, a Bayesian inference model is constructed for the target problem.
For example, Fig. \ref{fig:overview}(a) depicts a Bayesian network with two evidence nodes and two events, each having a uniform prior.
The likelihood probabilities are first modeled on the training data.
Then, the likelihoods are quantized to an adequate precision level without compromising the performance of the Bayesian model, as shown in Fig. \ref{fig:overview}(b).
These quantized likelihoods are mapped to multiple FeFET states corresponding to specifically defined $I_{DS}$s, associated with specific writing configurations for programming the FeFETs.
During the inference, discretized evidence values of the test samples activate corresponding crossbar columns.
The stored likelihoods are accumulated along each row, as shown in Fig. \ref{fig:overview}(c).
In this way, the posterior probabilities for each event are yielded as the crossbar outputs
without extra calculation circuitry.
These posteriors are sent to the sensing module to make the final prediction, completing one inference operation.

\vspace{-2ex}
\subsection{FeFET Crossbar Array Design and Operation}
\label{subsec:array}
\vspace{-0.5ex}

Fig.~\ref{fig:array} shows the proposed crossbar design of FeBiM, which includes the core FeFET array, row driver, write/input buffer, sensing module and peripheral circuitry.
In each array row,
the drain nodes of FeFETs connect to a  wordline ($WL$), and the source nodes are grounded to the sourceline ($ScL$).
The FeFETs within the same column share a common bitline ($BL$) at their gates.
We utilize a concise and scalable winner-take-all (WTA) circuit design \cite{liu2022cosime} in the sensing module to detect the $WL$ with the maximum current.

During write, the $WL$ and $ScL$ associated with the target row are grounded, and a 4V write voltage $V_{w}$ with corresponding pulse number is applied to the gate of target FeFETs, programming the designated quantized prior or likelihood probabilities.
To inhibit write disturbance, we apply a half bias $V_{w}/2$ scheme to the $WL$s and $ScL$s of unselected rows \cite{ni2018write}.
As indicated in Fig. \ref{fig:array}, 
in  orange/blue, 
the quantized prior/likelihood probabilities of a Bayesian model with $n$ evidence nodes (each evidence value is quantized to $m$ levels) and $k$ events are programmed into the corresponding sections of the crossbar.
During inference, the prior column is activated with $V_{on}$=0.5V on $BL_{1}$.
One column of each likelihood block is activated according to the input evidence value on $BL$s, and other unselected columns are inhibited with $V_{off}$=-0.5V.
The activated FeFET cells' $I_{DS}$  accumulate along each $WL$ as $I_{WL}$,  representing the calculated posteriors (denoted in green).
These $I_{WL}$s are then 
input to the WTA circuit (i.e., $I_{CM}$s) by the current mirrors.
The WTA circuit identifies the $WL$ with the maximum current, corresponding to the event with the highest posterior, 
and outputs a one-hot current result 
as the final inference decision 
(denoted with purple).

\vspace{-1.5ex}
\subsection{Probability Quantization and Mapping}
\label{subsec:mapping}
\vspace{-0.5ex}

To implement Eq.~\eqref{equ:bayes3} in our crossbar design, we adopt logarithmic computing.
In logarithmic domain, Eq.~\eqref{equ:bayes3} can be rewritten as:
\vspace{-1ex}
\begin{equation}
\label{equ:bayes_log}
\log P(A|B) \propto \log P(A) + \sum_{i=1}^{n} \log P(B_{i}|A)
\vspace{-1ex}
\end{equation}
This formula naturally aligns with our crossbar's computational behavior, 
as described in Sec. \ref{subsec:array}.
We initially convert the original probabilities into logarithmic values, then truncate very small probabilities to manage   
quantization precision efficiently. 
The quantization process involves two steps:
discretizing evidence values   to designated levels (corresponding with $m$ $BL$s in each likelihood block, as described in Sec. \ref{subsec:array}), and
quantizing the   logarithmic likelihoods
corresponding to the discretized evidence values 
with designated precision.
After quantization, we apply column normalization to the likelihoods corresponding to the same evidence value (i.e., the likelihoods stored in the same column) and priors:
\vspace{-1ex}
\begin{equation}
\label{equ:colnorm}
\begin{split}
P^{\prime}(A) &= \log P(A) + (1 - \max \log P(A)) \\
P^{\prime}(B_{i}=b|A) &= \log P(B_{i}=b|A) + (1 - \max \log P(B_{i}=b|A))
\end{split}
\vspace{-1ex}
\end{equation}
where each column of the normalized probabilities is added with a constant, with their maximum values scaled to 1.
This normalization enhances the differences among posteriors of multiple events 
without altering their order of magnitude, 
thus mitigating the accuracy degradation after quantization:
\vspace{-1ex}
\begin{equation}
\label{equ:colnorm_inf}
    \hat{A} = \arg \max_{A} \log P(A|B) = \arg \max_{A} P^{\prime}(A|B)
\vspace{-1ex}
\end{equation}
Finally, the normalized logarithmic probabilities $P^{\prime}$ are linearly mapped to discrete FeFET states with corresponding $I_{DS}$ values and
respective FeFET write configuration.
As illustrated in Fig. \ref{fig:mapping}(a), original probabilities $P$ (denoted with blue) are truncated (i.e., $P$<0.1 replaced with 0.1), converted to logarithmic values, and normalized to $P^{\prime}$.
$P^{\prime}$ is uniformly quantized to 10 levels, which are then linearly mapped to FeFET $I_{DS}$ values from 0.1 to 1.0 $\mu$A (denoted in orange).
Fig. \ref{fig:mapping}(b) shows the required gate pulse number to program FeFET into the designated state with required $I_{DS}$. 
For each discrete FeFET state associated with a quantized probability, the corresponding write configuration for programming the state into FeFET is determined thereof.

\vspace{-1ex}
\section{Validation and Evaluation}
\label{sec:eval}
\vspace{-1ex}

In this section, we investigate the functionality and scalability of FeBiM, and  benchmark its 
performance using a representative Bayesian classification task.
The results 
demonstrate
significant improvement in density and efficiency compared to the state-of-the-art designs.

\vspace{-2ex}
\subsection{Functionality and Performance}
\label{subsec:scalability}
\vspace{-1ex}

\begin{figure}[h]
\centering
\includegraphics[width=0.98\columnwidth]{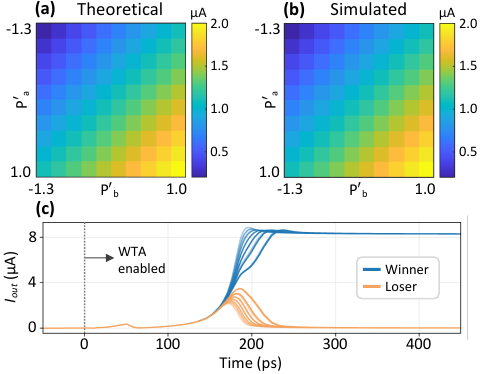}
\vspace{-3ex}
\caption{
When varying $P^{\prime}$ stored in two FeFETs, the (a) calculated $I_{WL}$ matches the (b) simulated $I_{WL}$ precisely.
(c) Transient simulation of the WTA circuit validates its function.}
\label{fig:veri}
\vspace{-5ex}
\end{figure}

We first verify the correct implementation of Eq. \eqref{equ:bayes_log} in the proposed crossbar design and the functionality of the WTA circuit.
In our SPECTRE simulations, the experimentally calibrated Preisach model \cite{ni2018circuit} is adopted for FeFETs, and
the 45nm PTM model \cite{vattikonda2006modeling} is utilized for MOSFETs with TT process corner at 27$^{\circ}$C and minimum sizes.
To begin with, we validate the posterior calculation with two likelihood factors in the logarithmic domain. 
Two FeFET cells $F_{a}$ and $F_{b}$ in the same row $WL_{i}$ are programmed with respective write configurations, storing corresponding $P^{\prime}_{a}$ and $P^{\prime}_{b}$, as depicted in Fig. \ref{fig:mapping}.
By varying $P^{\prime}_{a}$ and $P^{\prime}_{b}$, Fig.~\ref{fig:veri}(a) shows the theoretical $I_{WLi}$ values calculated from cell $I_{DS}$ values,  exactly matching the $I_{WLi}$ values obtained from circuit calculations shown in Fig.~\ref{fig:veri}(b).
We then validate the functionality of the WTA circuit.
Two separate $WL$s are connected to the WTA circuit, with $I_{WL1}$ and $I_{WL2}$ varying from 0.2$\mu$A and 2.0$\mu$A, respectively.
Fig.~\ref{fig:veri}(c) shows the  WTA circuit's output, where the winner results (i.e., $I_{WTA1}$ when $I_{WL1}>I_{WL2}$ and $I_{WTA2}$ when $I_{WL2}>I_{WL1}$) are clearly distinguishable from the loser results in less than 300ps.

\begin{figure}[!t]
\centering
\vspace{-1.5ex}
\includegraphics[width=0.96\columnwidth]{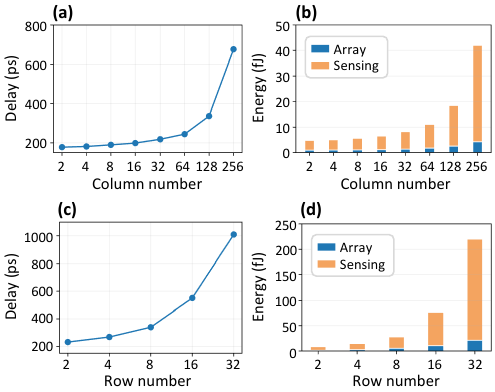}
\vspace{-3ex}
\caption{
The (a) inference delay and (b) energy consumption of FeBiM with 2 rows and a growing number of columns.
The (c) inference delay and (d) energy consumption of FeBiM with 32 columns and a growing number of rows.
}
\label{fig:scalability}
\vspace{-4.5ex}
\end{figure}

To evaluate the scalability and circuit-level performance of FeBiM, we test arrays with increasing numbers of rows and columns and measure the inference delay and energy consumption, as shown in Fig.~\ref{fig:scalability}.
In these tests, all $BL$s are activated.
The inference delay is measured as the time required to identify the winner output of the WTA circuit in the worst cases (i.e., minimum gap between adjacent $I_{WL}$ values) after activating  $BL$s.
The inference energy includes the consumption of the crossbar array and the sensing module.
The array part consists of the power dissipation in $WL$ drivers
and $BL$ drivers.
The sensing part includes the power consumption in current mirrors and the WTA circuit 
during inference.
As array size increases, longer stabilization time leads to larger delay and
higher power dissipation in both the crossbar and the sensing module.

\vspace{-1.5ex}
\subsection{Application Benchmarking}
\label{subsec:benchmark}
\vspace{-0.5ex}

A Bayesian classifier, often used in tasks such as spam detection or sentiment analysis, applies Bayesian inference to assign
a sample  to classes based on the highest posterior probability
\cite{wang2010don}.
As described in Eq.~\eqref{equ:bayes2}, the posterior class probabilities ($P(A|B)$) are estimated based on the likelihoods of observed features (evidence values) given each class ($P(B|A)$) and the prior probabilities of the classes ($P(A)$).
To simplify model construction and enhance likelihood computation, the Gaussian naive Bayesian classifier (GNBC), as a specific type of Bayesian classifier designed for continuous data, assumes
conditional independence of features given the class label (Eq.\ref{equ:bayes3}) and  a Gaussian distribution for each feature:
it estimates the mean and variance of each feature for each class and uses the Gaussian probability density function to compute the likelihood of a data point belonging to a class.
Despite its simplified assumptions, GNBC typically performs well in practice  \cite{rish2001empirical}.

\begin{table*}[!t]
\centering
\caption{Comparison between FeBiM and other Bayesian inference implementations with emerging devices.}
\vspace{-3ex}
\label{tab:compare}
\resizebox{\textwidth}{!}{%
\begin{tabular}{ccccccccccc}

    \hline\hline
    \multicolumn{1}{c|}{Reference} &
      \multicolumn{1}{c|}{Technology} &
      \multicolumn{1}{c|}{\begin{tabular}[c]{@{}c@{}} Device \\ usage \end{tabular}} &
      \multicolumn{1}{c|}{\begin{tabular}[c]{@{}c@{}} Device \\ configuration \end{tabular}} &
      \multicolumn{1}{c|}{\begin{tabular}[c]{@{}c@{}} Probability \\ storage unit \end{tabular}} &
      \multicolumn{1}{c|}{Calculation circuitry} &
      \multicolumn{1}{c|}{Sensing circuitry} &
      \multicolumn{1}{c|}{\begin{tabular}[c]{@{}c@{}} Speed \\ clk./inf. \end{tabular}} &
      \multicolumn{1}{c|}{\begin{tabular}[c]{@{}c@{}} Storage density \\ Mb/mm$^{2}$ \end{tabular}} &
      \multicolumn{1}{c|}{\begin{tabular}[c]{@{}c@{}} Computing density \\ MO/mm$^{2}$ \end{tabular}} &
      \begin{tabular}[c]{@{}c@{}} Efficiency \\ TOPS/W \end{tabular} \\ \hline
    \multicolumn{1}{c|}{\cite{vodenicarevic2017low}} &
      \multicolumn{1}{c|}{MTJ} &
      \multicolumn{1}{c|}{RNG} &
      \multicolumn{1}{c|}{SLC} &
      \multicolumn{1}{c|}{\textbackslash$^{\ast}$} &
      \multicolumn{1}{c|}{\begin{tabular}[c]{@{}c@{}} RNG, logic gate, compa-\\rator, Muller C-element \end{tabular}} &
      \multicolumn{1}{c|}{PCSA} &
      \multicolumn{1}{c|}{2000} &
      \multicolumn{1}{c|}{\textbackslash$^{\ast}$} &
      \multicolumn{1}{c|}{0.23} &
      0.013
       \\ \hline
    \multicolumn{1}{c|}{\cite{zheng2022hardware}} &
      \multicolumn{1}{c|}{Memtransistor} &
      \multicolumn{1}{c|}{RNG} &
      \multicolumn{1}{c|}{SLC} &
      \multicolumn{1}{c|}{\textbackslash$^{\ast}$} &
      \multicolumn{1}{c|}{RNG, logic gate} &
      \multicolumn{1}{c|}{Inverting amplifier} &
      \multicolumn{1}{c|}{200} &
      \multicolumn{1}{c|}{\textbackslash$^{\ast}$} &
      \multicolumn{1}{c|}{0.033} &
      0.0025
      \\ \hline
    \multicolumn{1}{c|}{\cite{harabi2023memristor}} &
      \multicolumn{1}{c|}{Memristor} &
      \multicolumn{1}{c|}{Memory} &
      \multicolumn{1}{c|}{SLC} &
      \multicolumn{1}{c|}{8$\times$2T2R cells$^{\dagger}$} &
      \multicolumn{1}{c|}{LFSR, comparator} &
      \multicolumn{1}{c|}{PCSA} &
      \multicolumn{1}{c|}{1$\sim$255$^{\ddagger}$} &
      \multicolumn{1}{c|}{2.47$^{\wedge}$} &
      \multicolumn{1}{c|}{0.034$^{\wedge}$} &
      2.14$\sim$13.39$^{\ddagger}$ 
      \\ \hline
    \multicolumn{1}{c|}{This work} &
      \multicolumn{1}{c|}{FeFET} &
      \multicolumn{1}{c|}{Memory} &
      \multicolumn{1}{c|}{MLC} &
      \multicolumn{1}{c|}{1F} &
      \multicolumn{1}{c|}{\textbackslash$^{\diamond}$} &
      \multicolumn{1}{c|}{WTA circuit} &
      \multicolumn{1}{c|}{1} &
      \multicolumn{1}{c|}{26.32} &
      \multicolumn{1}{c|}{0.69} &
      581.40 \\ \hline\hline
    \multicolumn{11}{l}{
      \begin{tabular}[c]{@{}l@{}} $\ast$: Not mentioned. $\diamond$: Not required. $\dagger$: For 8-bit quantized likelihoods. $\ddagger$: Depending on the operation scheme. $\wedge$: Cell area data from \cite{liu202033}, scaled to 45nm. Abbreviations. RNG: random number \\[-3pt] generator. PCSA: pre-charge sense amplifiers. LFSR: linear feedback shift register. clk./inf.:clock cycles per inference. MO: million operations. TOPS/W: trillion operations per second per watt. \end{tabular}}
\vspace{-4ex}
\end{tabular}
}
\end{table*}

\begin{figure}[!t]
\centering
\vspace{-2ex}
\includegraphics[width=0.96\columnwidth]{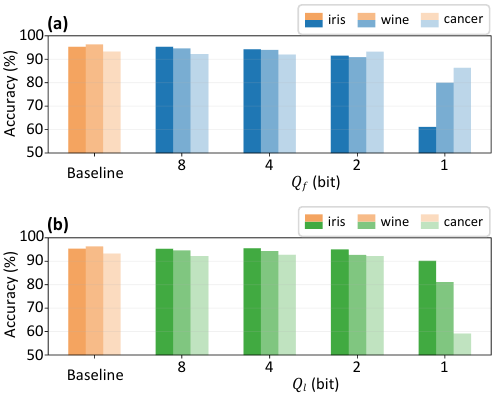}
\vspace{-3ex}
\caption{The inference accuracy of Gaussian Bayesian classifier on different datasets with respect to (a) varying feature quantization levels with 8-bit quantized likelihoods, and (b) varying likelihood quantization levels with 8-bit quantized features, when compared to the software baseline.
}
\label{fig:quantization}
\vspace{-5ex}
\end{figure}

Following the workflow of FeBiM in Fig.~\ref{fig:overview}, we 
construct GNBC models using Python Scikit-learn \cite{scikit-learn} and train them on three datasets, $iris$, $wine$ and $cancer$.
The test/train ratio is 0.7, and the number of training-inference epochs is set to 100 for each dataset to determine the average inference accuracy.
Fig.~\ref{fig:quantization} shows the software-based inference accuracy using 64-bit floating point precision. 
Then, we apply our proposed mapping scheme to convert these probabilities to logarithmic values for quantization and normalization.
Even with the feature quantization precision ($Q_{f}$, i.e., the levels of discretized evidence values as described in Sec. \ref{subsec:mapping}) or likelihood quantization precision ($Q_{l}$) reduced to as low as 2-bit, 
GNBCs display a negligible drop in inference accuracy on different datasets
compared to the software baseline.
These results indicate the effectiveness of our proposed mapping scheme in enabling 
high-performance Bayesian inference with limited precision.

\begin{figure}[!t]
\centering
\includegraphics[width=\columnwidth]{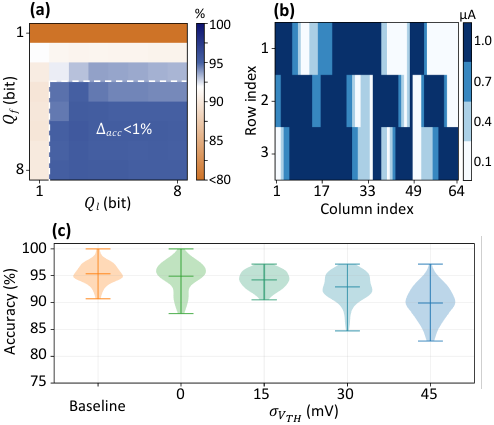}
\vspace{-6ex}
\caption{
(a) Average inference accuracy of $iris$-GNBC with different quantization precision.
(b) The states of FeFETs in a 3$\times$64 crossbar implementing the $iris$-GNBC with $Q_{f}$=4bit and $Q_{l}$=2bit.
(c) The inference accuracy distribution of circuit-implemented Bayesian classifier under different FeFET device variation levels as compared with the software baseline.
}
\label{fig:implementation}
\vspace{-4ex}
\end{figure}

We then implement the GNBC trained on $iris$ on our proposed FeFET-based crossbar architecture.
Fig.~\ref{fig:implementation}(a) shows the average inference accuracy of $iris$-GNBC with feature and likelihood quantization precisions varying from 1 to 8-bit.
The quantization region with $\Delta_{acc}<1\%$ is highlighted, where $\Delta_{acc}$ is the inference accuracy loss when compared to the software baseline.
We choose $Q_{f}$=4bit and $Q_{l}$=2bit as the optimal quantization precision, achieving an inference accuracy of 94.64\% with no significant gain observed at higher $Q_{f}$/$Q_{l}$ precision levels.
The quantized likelihoods are mapped to discrete FeFET $I_{DS}$ and 
programmed into the crossbar.
The $I_{DS}$ states of  FeFETs in the programmed crossbar storing the quantized likelihoods are shown in Fig.~\ref{fig:implementation}(b).
The array consists of 3 $WL$s for 3 classes and 64 $BL$s for 4 4-bit features.
Notably, due to the equal sample numbers in each class of $iris$, the prior is uniform and the prior block is omitted in Fig.~\ref{fig:implementation}(b) without affecting the posterior calculation and classification results.

To evaluate the inference accuracy of the hardware-implemented GNBC, we quantize the features of test samples as input for the crossbar.
As described in Sec. \ref{subsec:array}, each quantized feature value activates a corresponding $BL$, and the $I_{DS}$ of activated FeFETs are accumulated along each $ WL$.
The $WL$ with maximum accumulated current is then detected by the WTA circuit as the predicted class of the test sample.
The predictions of the in-memory GNBC are compared against the true labels of the test samples to calculate the classification accuracy.
The number of the training-inference epochs of the in-memory GNBC is set as 100.
The classification accuracy distribution of the circuit-implemented GNBC shows negligible degradation compared to the software baseline, as shown in Fig~\ref{fig:implementation}(c).
We further investigate the robustness of FeBiM using Monte Carlo simulations with varying levels of FeFET $V_{TH}$ variation.
Compared to the baseline, 
the mean accuracy drop is just  $\sim$5$\%$ at $\sigma_{V_{TH}}$=45mV.
Considering the experimental variation of FeFET device, such as \cite{soliman2023first} at 38mV, these results indicate the robustness and reliability of FeBiM in 
high-performance Bayesian classification.

In Table \ref{tab:compare}, we benchmark FeBiM's performance metrics in the context of $iris$-GNBC against other representative NVM-based Bayesian inference implementations.
In light of a 2FeFET per cell IMC design at 45nm node \cite{yin2020fecam}, we lay out a 2$\times$2 FeFET array and estimate a cell area of 0.076um$^{2}$.
The array density reaches 26.32 Mb/mm$^{2}$ when storing 2 bits per cell.
The average energy dissipation per inference is 17.20fJ, 
yielding a computing efficiency of  581.40 TOPS/W.
As a result, FeBiM outperforms the state-of-the-art memristor-based Bayesian machine by 10.7$\times$ in storage density and 43.4$\times$ in efficiency.
Compared to the RNG-based implementations, the computing density of FeBiM is improved by more than 3.0$\times$.
\section{Conclusion}
\label{sec:conclusion}
\vspace{-0.5ex}

In this paper, we introduce FeBiM, a highly compact and efficient in-memory Bayesian inference engine 
utilizing multi-bit FeFETs.
Our novel mapping scheme effectively  encodes the trained probabilities of a Bayesian inference model within a compact FeFET-based crossbar, enabling natural logarithmic in-memory Bayesian inference and facilitating more efficient and reliable computations.
When applied to a representative Bayesian classification task, FeBiM exhibited high inference accuracy and efficiency, maintaining robust performance despite  variations and limited precision.
As the first FeFET-based in-memory Bayesian inference engine, FeBiM achieved a remarkable  storage density of 26.32 Mb/mm$^{2}$ and a computing efficiency of 581.40 TOPS/W.
These results demonstrate significant  advancements over the state-of-the-art Bayesian machine, showcasing FeBiM's potential in enhancing a broad range of Bayesian inference applications.


\vspace{-1ex}
\begin{acks}
\vspace{-0.5ex}
This work was partially supported by National Key R\&D Program
of China (2022YFB4400300), NSFC (92164203, 62104213) and SGC Cooperation Project (Grant No. M-0612).
\end{acks}
\vspace{-1ex}

\bibliographystyle{ieeetr}
\bibliography{06_Selfbib.bib}

\begin{thebibliography}{10}

\bibitem{ielmini2018memory}
D.~Ielmini {\em et~al.}, ``In-memory computing with resistive switching devices,'' {\em Nature electronics}, vol.~1, no.~6, pp.~333--343, 2018.

\bibitem{shafiee2016isaac}
A.~Shafiee {\em et~al.}, ``Isaac: A convolutional neural network accelerator with in-situ analog arithmetic in crossbars,'' {\em ACM SIGARCH}, vol.~44, no.~3, pp.~14--26, 2016.

\bibitem{hu2021memory}
X.~S. Hu {\em et~al.}, ``In-memory computing with associative memories: A cross-layer perspective,'' in {\em 2021 IEEE IEDM}, pp.~25--2, 2021.

\bibitem{yan2023improving}
Z.~Yan {\em et~al.}, ``Improving realistic worst-case performance of nvcim dnn accelerators through training with right-censored gaussian noise,'' in {\em 2023 IEEE/ACM ICCAD}, pp.~1--9, 2023.

\bibitem{jung2022crossbar}
S.~Jung {\em et~al.}, ``A crossbar array of magnetoresistive memory devices for in-memory computing,'' {\em Nature}, vol.~601, no.~7892, pp.~211--216, 2022.

\bibitem{qayyum2020secure}
A.~Qayyum {\em et~al.}, ``Secure and robust machine learning for healthcare: A survey,'' {\em IEEE Reviews in Biomedical Engineering}, vol.~14, pp.~156--180, 2020.

\bibitem{yang2022unbox}
G.~Yang {\em et~al.}, ``Unbox the black-box for the medical explainable ai via multi-modal and multi-centre data fusion,'' {\em Information Fusion}, vol.~77, pp.~29--52, 2022.

\bibitem{ghahramani2015probabilistic}
Z.~Ghahramani, ``Probabilistic machine learning and artificial intelligence,'' {\em Nature}, vol.~521, no.~7553, pp.~452--459, 2015.

\bibitem{burkart2021survey}
N.~Burkart {\em et~al.}, ``A survey on the explainability of supervised machine learning,'' {\em Journal of Artificial Intelligence Research}, vol.~70, pp.~245--317, 2021.

\bibitem{smith2020massively}
R.~J. Smith {\em et~al.}, ``Massively parallel bayesian inference for transient gravitational-wave astronomy,'' {\em MNRAS}, vol.~498, no.~3, pp.~4492--4502, 2020.

\bibitem{talbot2019parallelized}
C.~Talbot {\em et~al.}, ``Parallelized inference for gravitational-wave astronomy,'' {\em Physical Review D}, vol.~100, no.~4, p.~043030, 2019.

\bibitem{awano2020bynqnet}
H.~Awano {\em et~al.}, ``Bynqnet: Bayesian neural network with quadratic activations for sampling-free uncertainty estimation on fpga,'' in {\em 2020 IEEE DATE}, pp.~1402--1407, IEEE, 2020.

\bibitem{vodenicarevic2017low}
D.~Vodenicarevic {\em et~al.}, ``Low-energy truly random number generation with superparamagnetic tunnel junctions for unconventional computing,'' {\em PRL}, vol.~8, no.~5, p.~054045, 2017.

\bibitem{zheng2022hardware}
Y.~Zheng {\em et~al.}, ``Hardware implementation of bayesian network based on two-dimensional memtransistors,'' {\em Nature communications}, vol.~13, no.~1, p.~5578, 2022.

\bibitem{faria2018implementing}
R.~Faria {\em et~al.}, ``Implementing bayesian networks with embedded stochastic mram,'' {\em AIP Advances}, vol.~8, no.~4, 2018.

\bibitem{harabi2023memristor}
K.-E. Harabi {\em et~al.}, ``A memristor-based bayesian machine,'' {\em Nature Electronics}, vol.~6, no.~1, pp.~52--63, 2023.

\bibitem{muller2012ferroelectricity}
J.~M{\"u}ller {\em et~al.}, ``Ferroelectricity in hfo 2 enables nonvolatile data storage in 28 nm hkmg,'' in {\em 2012 IEEE Symposium on VLSI technology}, pp.~25--26, IEEE, 2012.

\bibitem{khan2020future}
A.~I. Khan {\em et~al.}, ``The future of ferroelectric field-effect transistor technology,'' {\em Nature Electronics}, vol.~3, no.~10, pp.~588--597, 2020.

\bibitem{yin2024ferroelectric}
X.~Yin {\em et~al.}, ``Ferroelectric compute-in-memory annealer for combinatorial optimization problems,'' {\em Nature Communications}, vol.~15, no.~1, p.~2419, 2024.

\bibitem{dunkel2017fefet}
S.~D{\"u}nkel {\em et~al.}, ``A fefet based super-low-power ultra-fast embedded nvm technology for 22nm fdsoi and beyond,'' in {\em 2017 IEEE IEDM}, pp.~19--7, IEEE, 2017.

\bibitem{ni2019ferroelectric}
K.~Ni {\em et~al.}, ``Ferroelectric ternary content-addressable memory for one-shot learning,'' {\em Nature Electronics}, vol.~2, no.~11, pp.~521--529, 2019.

\bibitem{yin2022ferroelectric}
X.~Yin {\em et~al.}, ``Ferroelectric ternary content addressable memories for energy-efficient associative search,'' {\em IEEE TCAD}, vol.~42, no.~4, pp.~1099--1112, 2022.

\bibitem{li2020scalable}
C.~Li {\em et~al.}, ``A scalable design of multi-bit ferroelectric content addressable memory for data-centric computing,'' in {\em 2020 IEEE IEDM}, pp.~29--3, IEEE, 2020.

\bibitem{shou2023see}
S.~Shou {\em et~al.}, ``See-mcam: Scalable multi-bit fefet content addressable memories for energy efficient associative search,'' in {\em 2023 IEEE/ACM ICCAD}, pp.~1--9, IEEE, 2023.

\bibitem{soliman2023first}
T.~Soliman {\em et~al.}, ``First demonstration of in-memory computing crossbar using multi-level cell {FeFET},'' {\em Nature Communications}, vol.~14, no.~1, p.~6348, 2023.

\bibitem{yin2023ultracompact}
X.~Yin {\em et~al.}, ``An ultracompact single-ferroelectric field-effect transistor binary and multibit associative search engine,'' {\em Advanced Intelligent Systems}, vol.~5, no.~7, p.~2200428, 2023.

\bibitem{yin2021deep}
X.~Yin {\em et~al.}, ``Deep random forest with ferroelectric analog content addressable memory,'' {\em arXiv preprint arXiv:2110.02495}, 2021.

\bibitem{box2011bayesian}
G.~E. Box {\em et~al.}, {\em Bayesian inference in statistical analysis}.
\newblock John Wiley \& Sons, 2011.

\bibitem{nikovski2000constructing}
D.~Nikovski, ``Constructing bayesian networks for medical diagnosis from incomplete and partially correct statistics,'' {\em IEEE TKDE}, vol.~12, no.~4, pp.~509--516, 2000.

\bibitem{trimmer2011decision}
P.~C. Trimmer {\em et~al.}, ``Decision-making under uncertainty: biases and bayesians,'' {\em Animal cognition}, vol.~14, pp.~465--476, 2011.

\bibitem{lowd2005naive}
D.~Lowd {\em et~al.}, ``Naive bayes models for probability estimation,'' in {\em ICML}, pp.~529--536, 2005.

\bibitem{ko20203mm}
G.~G. Ko {\em et~al.}, ``A 3mm 2 programmable bayesian inference accelerator for unsupervised machine perception using parallel gibbs sampling in 16nm,'' in {\em 2020 IEEE Symposium on VLSI Circuits}, IEEE, 2020.

\bibitem{liu2022cosime}
C.-K. Liu {\em et~al.}, ``Cosime: Fefet based associative memory for in-memory cosine similarity search,'' in {\em 2022 IEEE/ACM ICCAD}, pp.~1--9, 2022.

\bibitem{ni2018write}
K.~Ni {\em et~al.}, ``Write disturb in ferroelectric fets and its implication for 1t-fefet and memory arrays,'' {\em IEEE EDL}, vol.~39, no.~11, pp.~1656--1659, 2018.

\bibitem{ni2018circuit}
K.~Ni {\em et~al.}, ``A circuit compatible accurate compact model for ferroelectric-fets,'' in {\em 2018 IEEE Symposium on VLSI Technology}, pp.~131--132, IEEE, 2018.

\bibitem{vattikonda2006modeling}
R.~Vattikonda {\em et~al.}, ``Modeling and minimization of pmos nbti effect for robust nanometer design,'' in {\em 2006 ACM/IEEE DAC}, pp.~1047--1052, 2006.

\bibitem{wang2010don}
A.~H. Wang, ``Don't follow me: Spam detection in twitter,'' in {\em 2010 IEEE SECRYPT}, pp.~1--10, IEEE, 2010.

\bibitem{rish2001empirical}
I.~Rish {\em et~al.}, ``An empirical study of the naive bayes classifier,'' in {\em IJCAI}, vol.~3, pp.~41--46, 2001.

\bibitem{liu202033}
Q.~Liu {\em et~al.}, ``33.2 a fully integrated analog reram based 78.4 tops/w compute-in-memory chip with fully parallel mac computing,'' in {\em 2020 IEEE ISSCC}, pp.~500--502, IEEE, 2020.

\bibitem{scikit-learn}
F.~Pedregosa {\em et~al.}, ``Scikit-learn: Machine learning in {P}ython,'' {\em the Journal of Machine Learning Research}, vol.~12, pp.~2825--2830, 2011.

\bibitem{yin2020fecam}
X.~Yin {\em et~al.}, ``Fecam: A universal compact digital and analog content addressable memory using ferroelectric,'' {\em IEEE TED}, vol.~67, no.~7, pp.~2785--2792, 2020.

\end{thebibliography}

\end{document}